\documentclass{article}

\usepackage[preprint]{neurips_2026}
\usepackage{neurips_2026}
\usepackage{adjustbox}
\usepackage[utf8]{inputenc} % allow utf-8 input
\usepackage[T1]{fontenc}    % use 8-bit T1 fonts
\usepackage{hyperref}       % hyperlinks
\usepackage{url}            % simple URL typesetting
\usepackage{booktabs}       % professional-quality tables
\usepackage{amsfonts}       % blackboard math symbols
\usepackage{nicefrac}       % compact symbols for 1/2, etc.
\usepackage{microtype}      % microtypography
\usepackage{xcolor}         % colors
\usepackage{algorithm}
\usepackage{algpseudocode}

\usepackage{orcidlink}
\usepackage{multirow}
\usepackage{graphicx}
\usepackage{amsmath}
\definecolor{citecolor}{HTML}{0071bc}
\definecolor{paleplum}{rgb}{0.8, 0.6, 0.8}
\hypersetup{
  colorlinks,
  citecolor=citecolor,
  linkcolor=red
}

\title{Dynamic Execution Commitment of Vision-Language-Action Models}

\author{%
    \parbox{\textwidth}{\centering
        \textbf{Feng Chen\textsuperscript{1}\thanks{Equal contribution. $\dagger$  Corresponding author.}} ~~
        \textbf{Xianghui Wang\textsuperscript{2$*$}} ~~
        \textbf{Yuxuan Chen\textsuperscript{3}} ~~
        \textbf{Boying Li\textsuperscript{4$\dagger$}} ~~
        \textbf{Yefei He\textsuperscript{5}} ~~
    }\\[6pt]
    \parbox{\textwidth}{\centering
        \textbf{Zeyu Zhang\textsuperscript{5}}
        \textbf{Yicheng Wu\textsuperscript{6$\dagger$}}
    }\\[10pt]
    \parbox{\textwidth}{\centering
        \textsuperscript{1} University of Adelaide \hspace{0.5em}
        \textsuperscript{2} Sichuan University \hspace{0.5em}
        \textsuperscript{3} Shanghai Jiao Tong University
    }\\
    \parbox{\textwidth}{\centering
        \textsuperscript{4} Monash University \hspace{0.5em}
        \textsuperscript{5} Zhejiang University \hspace{0.5em}
        \textsuperscript{6} Imperial College London
    }
}

\begin{document}

\maketitle
\begin{abstract}
Vision-Language-Action (VLA) models predominantly adopt
\textbf{action chunking}, \textit{i.e.}, predicting a short sequence of
consecutive low-level actions in a single forward pass and committing them
to execution, to amortise the inference cost of large-scale backbones and
reduce per-step latency. However, determining how many predicted actions
to execute requires balancing success rate against inference efficiency,
a decision typically governed by fixed execution horizons tuned per task.
Such heuristics ignore the state-dependent nature of predictive reliability,
leading to brittle performance in dynamic or out-of-distribution settings.
In this paper, we introduce A$^3$, an Adaptive Action Acceptance mechanism
that reframes dynamic execution commitment as a self-speculative prefix
verification problem. A$^3$ first estimates action-wise consensus across
sampled trajectories, based on which it selects a representative draft and
prioritizes downstream verification. Specifically, it enforces:
(1) consensus-ordered conditional invariance, which validates low-consensus
actions by determining whether they remain consistent when re-decoded
conditioned on high-consensus actions; and
(2) prefix-closed sequential consistency, which accepts an action only if
it and all preceding actions pass verification, preventing execution from
advancing beyond the first failure. Consequently, the execution
horizon is determined as the longest prefix satisfying both conditional
invariance and sequential consistency. Experiments across diverse VLA
models and benchmarks demonstrate that A$^3$ eliminates the need for manual
horizon tuning while achieving a superior trade-off between execution
robustness and inference throughput. Our code is available at
\href{https://inceptionwang.github.io/A3/}{here}.

\end{abstract}

\section{Introduction}
Vision-Language-Action (VLA) models \cite{pi05,openvla,gr3} have emerged as a primary paradigm for generalizable embodied intelligence \cite{yu_survey_efficient_vla_2025,ma_survey_vla_embodied_2024}, mapping high-dimensional visual observations and natural language instructions directly to precise motor sequences. To mitigate the high computational overhead of large-scale vision-language backbones \cite{shao_survey_vlm_based_vla_2025}, modern VLA architectures increasingly adopt dual-system designs \cite{pi05} that decouple deliberative reasoning from reactive execution. A core efficiency strategy within these frameworks is action chunking, where an action expert (e.g., $\pi$-0.5 \cite{pi05} or GR00T \cite{gr00t}) generates a sequence of future actions in a single forward pass before interacting with the environment again. While substantially reducing inference frequency and improving computational utilization, the determination of the execution horizon, \textit{i.e.,} the number of actions actually committed before the next re-planning cycle, remains a critical yet underexplored design choice.

\begin{figure}
    \centering
    \includegraphics[width=1\linewidth]{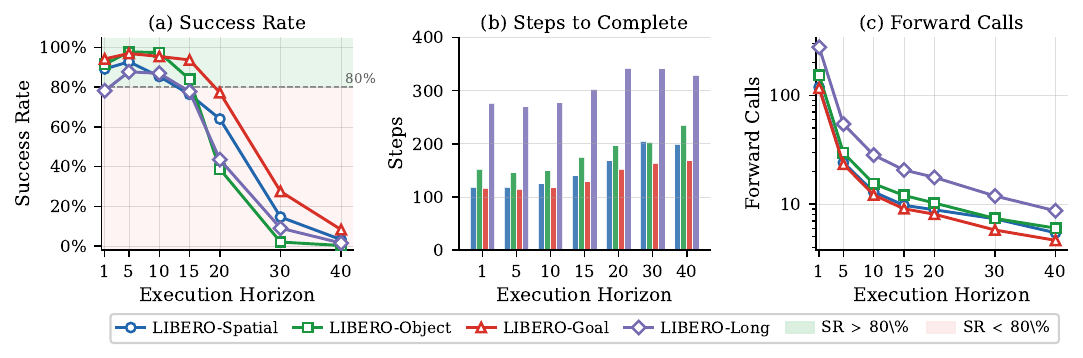}
    \caption{Performance analysis of $\pi$-0.5 under varying execution horizons on LIBERO benchmark \cite{libero}.
(a) Success rate first increases while then decreases as horizon increases, dropping below 80\% when horizon is larger than 15 for most suites.
(b) Completion step increases substantially with a larger horizon, as failed recoveries and compounding errors drive total steps up to $1.5\times$ higher than at horizon=1.
(c) Forward calls decrease monotonically with a longer horizon, revealing a fundamental trade-off between inference overhead and success rate.}
    \label{fig:horizon_analysis}
\end{figure}

In existing VLA systems, execution commitment is typically restricted to a predefined, static horizon selected a priori per task \cite{black2024pi_0}. This design implicitly assumes that the model's predictive reliability is uniform across the temporal rollout and invariant to environmental stochasticity. However, in open-ended embodied settings, observations are non-stationary and execution errors inevitably compound, rendering a fixed execution horizon substantially brittle~\cite{moh,everydayvla}. As illustrated in Figure~\ref{fig:horizon_analysis}, it reveals a fundamental trade-off between inference efficiency and task success rate under varying execution horizons: a smaller execution horizon requires more frequent model queries, imposing heavy inference overhead on real-time robot control, whereas a larger horizon reduces forward calls but commits to longer open-loop action sequences, resulting in severe success rate degradation as accumulated errors cannot be corrected by visual feedback.

This motivates the design of an adaptive mechanism that reduces unnecessary forward calls while preserving the reactivity needed to maintain high task success. While recent efforts have begun exploring adaptive execution, existing approaches can be broadly categorized into two paradigms, each with distinct limitations. \textit{Consistency-based methods}, such as Mixture of Horizons (MoH)~\cite{moh} and 
EveryDayVLA~\cite{everydayvla}, estimate reliability by measuring agreement across 
multiple prediction sources---either cross-horizon segment consensus or discrepancy 
between discrete and continuous action heads. However, these approaches either require 
additional training overhead or produce heuristic thresholds that are neither calibrated 
nor grounded in sequential execution feasibility. \textit{Attention-based methods}, such as AutoHorizon~\cite{autohorizon}, leverage self-attention weights as a proxy \cite{gu2024attention,kang2025see} for predictive confidence, but attention patterns do not directly reflect the physical feasibility of committing an action sequence in an open-loop manner. As a result, none of these methods provides a principled mechanism that jointly accounts for model-level self-consistency and execution-level sequential feasibility in continuous action spaces.

To resolve this tension, we introduce Adaptive Action Acceptance (A$^3$), a mechanism that reframes dynamic execution commitment as a self-speculative prefix verification problem. Inspired by the principles of self-speculative decoding~\cite{ssd,leviathan2023fast}, A$^3$ treats the predicted action sequence as a draft and adaptively determines the execution horizon by assessing the model's internal logic consistency and sequential execution consistency, without auxiliary modules or retraining. We first compute a mode-aware trajectory consensus score of actions over multiple sampled rollouts. The goal of consensus scoring is to estimate, for each action, how stably the model predicts that action across independent rollouts - not in raw action space (where incremental errors compound), but in the space of induced trajectory states. This score serves as a self-consistency prior for representative draft selection and verification ordering, rather than a calibrated estimate of correctness. Then A$^3$ determines the execution window through a dual hierarchical verification pipeline. First, it enforces \textit{consensus-ordered conditional invariance}: lower-consensus actions are re-decoded conditioned on higher-consensus actions fixed as context; an action is accepted only if it remains invariant under this re-decoding, ensuring the speculative segment is consistent with the dominant predicted plan. Second, it imposes \textit{prefix-closed sequential consistency}: each action is re-decoded
conditioned on all its temporally preceding draft actions; an action is
accepted only if this re-decoding confirms invariance and every preceding
action has also passed verification, thereby terminating execution at the
first verification failure. The final execution horizon emerges as the longest prefix satisfying both constraints, resolved in a single parallel forward pass—adapting naturally to task difficulty and observation quality without any task-specific tuning. Our contributions can be summarized as:

 • We formulate adaptive execution commitment as a self-speculative
    prefix verification problem for continuous-action VLA policies. Unlike
    prior methods that estimate execution horizons from indirect proxy
    signals, our formulation actively re-decodes candidate actions under
    structured contexts and determines the horizon as the longest
    conditionally stable prefix.

    • We instantiate this formulation in A$^3$ through mode-aware
    trajectory consensus and dual hierarchical verification. Consensus in
    induced trajectory space selects a representative draft and determines
    the verification order, while consensus-ordered conditional invariance
    and prefix-closed sequential consistency jointly determine the committed
    action prefix.

• Extensive experiments across multiple VLA backbones, simulation
    benchmarks, and real-world manipulation tasks demonstrate that A$^3$
    improves the trade-off between task success and inference efficiency
    without task-specific horizon tuning.

\section{Related Work}

\noindent \textbf{Vision-language-action model.} VLA models \cite{spatialvla_rss2025,taskcentric_latent_actions_rss2025} aim to learn policies that map visual observations and language instructions \cite{bai2025qwen3,deng2025emerging} directly to continuous control actions \cite{open_x_embodiment_rtx_2023}. Early approaches \cite{openvla} primarily relied on autoregressive or behavior-cloning policies that predict actions step by step. 
More recently, dual-system VLA architectures \cite{pi05,black2024pi_0} have emerged, decoupling high-level multimodal reasoning \cite{bai2025qwen3,beyer2024paligemma} from low-level action generation. 
In these systems, a perception-language backbone \cite{bai2025qwen3} produces contextual representations, while a dedicated action expert is often implemented as a diffusion \cite{diffusion_policy_rss2023} or flow-matching head that predicts multiple future actions in parallel \cite{parallel_decoding_chunking_2025,freqpolicy_2025,act_rss2023}.
In practice, these systems often execute a predefined number of predicted actions before re-planning, with the execution horizon typically selected per task or benchmark \cite{calvin_ral2022,rlbench_ral2020,droid_rss2024}.
For instance, $\pi$-0.5 \cite{pi05} adopts different execution horizons across LIBERO subtasks \cite{libero} to balance performance and stability. 
Although effective under benchmark-specific settings, such task-dependent configuration relies on prior knowledge of task structure.

\noindent \textbf{Speculative decoding.} Speculative decoding \cite{leviathan2023fast,he2024zipar} was originally proposed to accelerate autoregressive language models by drafting multiple tokens and verifying them using a stronger or identical model. 
Subsequent work, including self-speculative decoding \cite{ssd,xia2024swift}, further improves efficiency by leveraging structured verification trees to enable parallel token prediction while preserving exact decoding correctness in discrete token spaces. 
For instance, SSD \cite{ssd} leverages self-drafting to eliminate auxiliary models, enabling dLLMs \cite{llada} to generate and verify multiple tokens per iteration \cite{ke2025explain}.
While structurally related \cite{specvla_2025}, our work addresses a fundamentally different objective. 
Speculative decoding aims to improve generation efficiency without altering model outputs. 
In contrast, we focus on execution commitment in VLA systems—determining how many predicted actions to commit before replanning under the current state. 
Instead of enforcing exact token matching, our framework models commitment as a consensus-ordered, prefix-closed verification process that explicitly accounts for uncertainty in continuous action spaces and prefix-wise sequential consistency.

\noindent \textbf{Adaptive execution and horizon selection.}  Prior work has explored adaptive horizons primarily through model-based control (MPC) \cite{holkar2010overview} or confidence-based heuristics \cite{kouvaritakis2016model,holkar2010overview}. In the VLA context,  EveryDayVLA ~\cite{everydayvla} utilizes the AdaHorizon algorithm to adjust windows based on disagreement between discrete and continuous action heads.  Similarly, NanoVLA \cite{chen2025nanovla} employs Long-Short Action Chunking to plan long but act short. 
 Mixture-of-Horizon (MoH) \cite{moh} rearranges action chunks into multiple segments with different horizons and fuses their predictions through a lightweight linear gating mechanism. Moreover, AutoHorizon \cite{autohorizon} depends on the attention weight as prediction confidence to infer the temporal limit of the model’s reliable forecasting capability. In contrast, A$^3$ does not rely on explicit dynamics models, auxiliary heads, or passive thresholds. Instead, it treats commitment as a self-consistent prefix verification problem where the horizon emerges from conditional invariance, providing a principled way to scale execution without task-specific tuning.

%\noindent \textbf{Model predictive control}
\section{Method}

\begin{figure}
    \centering
    \includegraphics[width=1\linewidth]{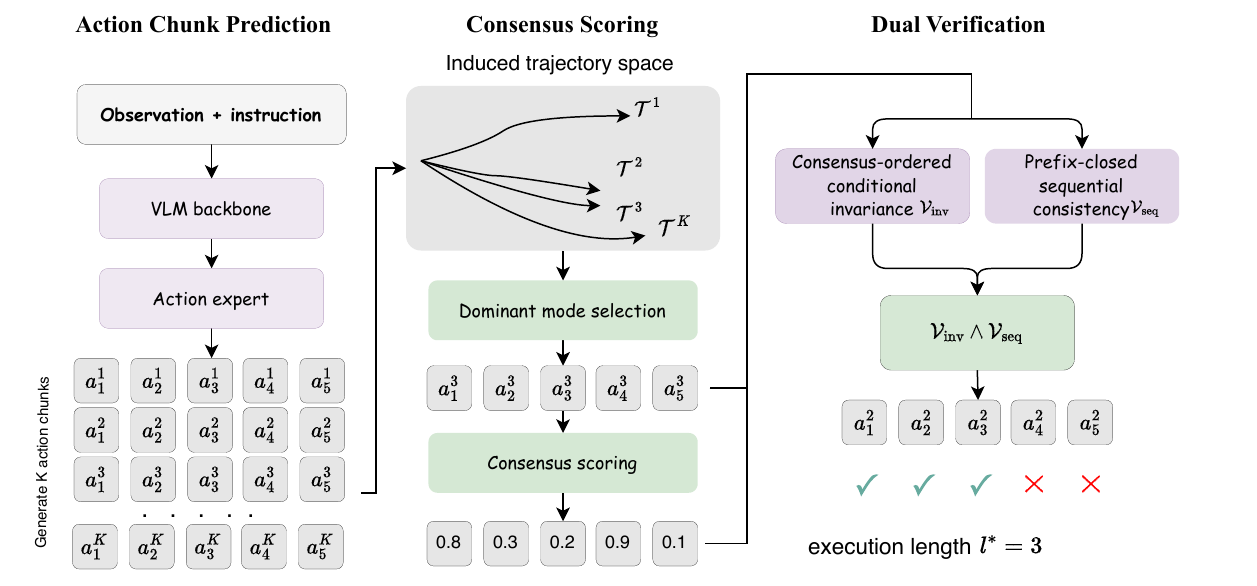}
    \caption{Overview of $\text{A}^3$. Given the current observation and instruction, the VLM backbone 
and action expert generate $K$ candidate action chunks. The chunks are mapped to induced 
trajectory states, from which the dominant mode is identified via clustering and its 
medoid selected as the primary draft; per-step consensus scores reflect 
the model's self-consistency at each action position. The draft then undergoes dual parallel 
verification in a single batched 
forward pass, yielding the execution horizon $l^*$ as the longest verified prefix. We also illustrate the implementation of dual verification tree in Figure \ref{fig:main}.}
    \label{fig: main1}
\end{figure}

As illustrated in Figure \ref{fig: main1}, the A$^3$ framework transforms static action commitment into an adaptive, state-conditioned verification process via a three-stage pipeline. First, A$^3$ generates $K$ candidate chunks in a single forward pass and then computes a mode-aware trajectory consensus score to select a primary draft sequence. The draft is then verified through a parallel tree governed by two hierarchical constraints: (1) \textit{Consensus-ordered Conditional Invariance}, which validates lower-consensus actions by judging whether they remain invariant when re-decoded conditioned on higher-consensus actions; and (2) \textit{Prefix-closed Sequential Consistency}, which re-decodes each
action conditioned on its preceding draft actions and excludes it from
execution if either the action itself or any of its predecessors fails
verification. Consequently, the final execution horizon $l^*$ is determined
as the longest prefix in which every action satisfies both conditional
invariance and sequential consistency.

\subsection{Execution Commitment Formulation}

Consider a VLA policy $\pi_\theta$ that, given the current state $s_t$ (including visual observations and language instructions), predicts a sequence of future actions of length $H$, written as
\begin{equation}
\mathbf{A}_t = \pi_\theta(s_t) =  \{ a_t, a_{t+1}, \dots, a_{t+H-1} \}.
\end{equation}

% Due to the non-stationary nature of embodied environments and the accumulation of prediction errors, committing to the entire sequence $\mathbf{A}_t$ in an open-loop fashion is often suboptimal. 
Due to the non-stationarity of embodied environments and the progressive accumulation of prediction errors, executing the full action sequence $\mathbf{A}_t$ in an open-loop manner is generally suboptimal. Instead, the system should adaptively determine an appropriate execution horizon $l$ to balance success rate against inference efficiency, as illustrated in Figure \ref{fig:horizon_analysis}.
We formalize execution commitment by selecting a prefix length $l \leq H$ and defining the committed action subsequence as
\begin{equation}
\mathbf{A}_t[1:l] = \{ a_t, \dots, a_{t+l-1} \},
\end{equation}
which is selected for execution before the next replanning step under the current state $s_t$.
% Formally, the objective is to determine
% \begin{equation}
% l^* = \max_{l \in \{1, \dots, T\}} \; \mathbb{I}\big( \mathcal{V}(\mathbf{A}_t^{1:l}, s_t) = 1 \big),
% \end{equation}
% where $\mathcal{V}(\cdot)$ denotes a verification function that evaluates whether the prefix satisfies execution constraints under the current context.
Formally, we determine the prefix length by solving the following constrained optimization problem:
\begin{equation}
l^* = \mathop{\arg\max}\limits_{l \in \{1, \dots, H\}} l
\quad
\text{s.t.} \quad
\prod_{\tau = t}^{t+l-1} \mathcal{V}(a_\tau, s_\tau) = 1.
\end{equation}
Here, $\mathcal{V}(a_\tau, s_\tau) \in \{0,1\}$ is a verification function that indicates whether the candidate action prefix satisfies the execution constraints under the current state $s_t$.

\subsection{Mode-aware Trajectory Consensus Scoring}

The primary challenge in assessing the consistency of an action sequence $\mathbf{A}_t$ is the intra-sequence visual open-loop vulnerability. Since VLA models typically output incremental displacements (e.g., $\Delta$-poses), small per-step prediction errors accumulate through temporal integration, leading to significant task-space drift. Traditional uncertainty metrics that focus on marginal action variance at each timestep~\cite{everydayvla} fail to capture this compounding physical reality.

\textbf{Induced trajectory states.} To account for error propagation, A$^3$ evaluates consistency in the space of induced trajectory states. We define the induced state at a future step $t+j$ as:
\begin{equation}
    \hat{s}_{t+j} = \mathcal{G}(s_t, \{a_t, \dots, a_{t+j-1}\}),
\end{equation}
where $\mathcal{G}$ denotes kinematic integration~\cite{montana1988kinematics} of incremental actions into a cumulative pose. A$^3$ probes the model's posterior by generating $K$ independent candidate sequences $\{ \mathbf{A}_t^{(1)}, \dots, \mathbf{A}_t^{(K)} \}$ in a single forward pass, yielding a set of induced trajectories $\{\mathcal{T}^{(1)}, \dots, \mathcal{T}^{(K)}\}$.

\textbf{Dominant mode selection.} Rather than measuring consensus against a global centroid, which conflates multi-modal but internally coherent futures with genuine instability, we identify the dominant trajectory mode via clustering. Since spatially similar rollouts may be temporally phase-shifted, we measure pairwise trajectory agreement using a locally time-aligned distance $d_j^{(k,k')} = \min_{|\Delta| \leq w} \| \hat{s}_{t+j}^{(k)} - \hat{s}_{t+j+\Delta}^{(k')} \|_2^2$, aggregated over steps as $D(k,k') = \sum_j d_j^{(k,k')}$. We cluster the $K$ trajectories under this distance and score each cluster $c$ by
\begin{equation}
    S_c = p_c \cdot \exp(-\mathrm{Disp}(c)/\tau),
\end{equation}
where $p_c = |c|/K$ is the cluster mass, $\mathrm{Disp}(c)$ is the mean pairwise distance within the cluster, and $\tau > 0$ is a temperature parameter controlling the compactness penalty. The dominant mode is then selected as $c^* = \arg\max_c S_c$, balancing trajectory coherence with population support.

\textbf{Consensus score of action.} Within dominant mode $c^*$, we select the medoid trajectory $k^* = \arg\min_{k \in c^*} \sum_{k' \in c^*} D(k,k')$ as the primary draft, and define the action consensus score as:
\begin{equation}
    \tilde{R}_j = p_{c^*} \cdot \exp\!\left( -\frac{1}{|c^*|} \sum_{k \in c^*} d_j^{(k, k^*)} \right),
\end{equation}
where $d_j^{(k,k^*)}$ is the time-aligned distance from sample $k$ to the medoid at step $j$. Higher $\tilde{R}_j$ indicates that rollouts concentrate around a stable dominant plan at that step.

Importantly, we do not interpret $\tilde{R}_j$ as a calibrated measure of correctness. Rather, it serves as a \textit{self-consistency prior}: it selects the primary draft and determines the order in which actions are subjected to downstream verification. Final execution commitment is decided by the subsequent conditional-invariance and prefix-consistency checks, rather than by consensus alone.

\subsection{Dual Hierarchical Verification}

Following the drafting phase, A$^3$ evaluates the draft's conditional stability through structured re-decoding. Unlike token-level matching in LLMs, our framework treats verification as a progressive anchoring process governed by two constraints designed for embodied consistency.

\textbf{Consensus-ordered conditional invariance.}
This constraint evaluates whether a speculative action remains stable when the model's highest-consensus predictions are fixed as context. For each candidate action $a_{t+i}$, we define a consensus context $\mathcal{C}_i^{\mathrm{inv}}$ consisting of all actions with higher consensus scores:
\begin{equation}
\mathcal{C}_i^{\mathrm{inv}} = \{a_{t+j} \mid \tilde{R}_j > \tilde{R}_i,\ j \in \{1, \dots, H\} \},
\end{equation}
and verify $a_{t+i}$ by re-decoding it conditioned on this context \cite{black2026real}:
\begin{equation}
\mathcal{V}_{\mathrm{inv}}(a_{t+i}, s_t \mid \mathcal{C}^{\mathrm{inv}}_i) = \mathbb{I}\!\left(\left|\pi_\theta(s_t \mid \mathcal{C}^{\mathrm{inv}}_i)[i] - a_{t+i}\right| < \delta \right),
\end{equation}
where $\pi_\theta(s_t \mid \mathcal{C}_i^{\mathrm{inv}})[i]$ is the re-decoded output at position $i$, and $\delta > 0$ is an acceptance threshold. An action is accepted only if the model re-confirms it after its higher-consensus counterparts are locked in, ensuring that acceptance is consistent with the dominant sampled plan rather than marginal prediction alone.

\textbf{Prefix-closed sequential consistency.} 
Robotic execution is a strictly causal and non-reversible process. A failure at action $j$ invalidates the environmental assumptions for all subsequent actions, as the resulting state would diverge significantly from the model's training distribution—a phenomenon known as covariate shift. To prevent the robot from skipping across unverified segments, we enforce prefix-closed sequential consistency. Under this constraint, an action $i$ is only valid if all preceding actions in the chain are treated as fixed context and the action remains stable under re-decoding.
For each action $i$, we define a sequential prefix context consisting of all preceding actions:
\begin{equation}
\mathcal{C}_i^{\mathrm{seq}} = \{ a_{t+j} \mid j < i \}.
\end{equation}
We then perform an analogous re-decoding procedure for $a_{t+i}$ under the sequential context and verify it as:
\begin{equation}
\mathcal{V}_{\mathrm{seq}}(a_{t+i}, s_t \mid \mathcal{C}_i^{\mathrm{seq}})
=
\mathbb{I}
\left(
\left|
\pi_\theta(s_t \mid \mathcal{C}_i^{\mathrm{seq}})[i]
-
a_{t+i}
\right|
<
\delta
\right).
\end{equation}

% \begin{equation}
%     \Phi_{seq}(i) = \bigwedge_{j=1}^i \Phi_{inv}(j)
% \end{equation}

%This logical conjunction ensures the verification is prefix-closed: if the chain breaks at step $i$, the entire execution window is truncated at $i-1$, regardless of the stability of later speculative segments.

\subsection{Emergent Horizon Determination}
These two constraints address complementary failure modes: conditional
invariance resolves ambiguity within the predicted plan, whereas sequential
consistency preserves the causal validity of its physical execution.
Hierarchical verification accepts only actions satisfying both conditions.
Therefore, the final execution commitment $l^*$ emerges as the global solution to the following constrained maximization problem:
\begin{equation}
\begin{gathered}
l^* = \mathop{\arg\max}\limits_{l \in \{1,\dots,H\}} \; l \\
\text{s.t.} \quad
\mathcal{V}_{\mathrm{inv}}(a_{t+l}, s_t \mid \mathcal{C}_l^{\mathrm{inv}}) = 1, \quad
\mathcal{V}_{\mathrm{seq}}(a_{t+l}, s_t \mid \mathcal{C}_l^{\mathrm{seq}}) = 1.
\end{gathered}
\end{equation}
% \begin{equation}
%     l^* = \max \big\{ l \in \{1, \dots, T\} \mid \Phi_{seq}(l) = 1 \big\}
% \end{equation}
To maintain real-time throughput, A$^3$ implements this progressive verification as a hierarchical tree over candidate prefixes. By exploiting the parallel processing of the action expert, the entire dual-stage verification for the prediction horizon $H$ is completed in a single parallel forward pass. %Under this formulation, $l^*$ is an emergent property of the VLA self-assessed certainty. In predictable phases (e.g., reaching), the model exhibits high consensus and $l^*$ expands toward $H$, maximizing efficiency. Conversely, in high-precision phases (e.g., insertion), the verification chain breaks early as uncertainty rises , causing $l^*$ to contract automatically. This training-free mechanism allows A$^3$ to serve as a principled safety boundary without fine-tuning.
More implementation details and the algorithm can be found in Sec \ref{sec detail}.

\section{Experiments}

\subsection{Experimental Setup}

% \textbf{Implementation details.} We implement A$^3$ on three dual-system VLA backbones: $\pi$-0, $\pi$-0.5, and GR00T. All model configurations are adopted from the RLinf codebase. The prediction horizon $T$ follows each backbone's default setting ($H=10$ for $\pi$-0/$\pi$-0.5, $H=8$ for GR00T). During group probing, we generate $K=8$ candidate chunks via parallel noise sampling in a single batched forward pass. For dual verification, the invariance constraint is applied independently on each action dimension, with the threshold $\delta$ defaulting to the per-dimension standard deviation used for output denormalization. All experiments are conducted on a single NVIDIA RTX 4090 GPU.

\noindent\textbf{Implementation details.} We implement A$^3$ on three VLA models including $\pi_{0}$ \cite{black2024pi_0}, $\pi_{0.5}$ \cite{pi05}, and GR00T \cite{gr00t}. All model configurations are adopted from the RLinf codebase \cite{rlinf}. The prediction horizon $H$ follows each backbone's default setting on the respective datasets. Specifically, we set $H=10$ for LIBERO \cite{libero}, $H=8$ for ManiSkill \cite{maniskill}, and $H=5$ for MetaWorld \cite{metaworld} when using $\pi_{0}$ and $\pi_{0.5}$, whereas $H=5$ is adopted for GR00T. During group probing, we generate $K=8$ candidate chunks via parallel noise sampling in a single batched forward pass. For dual verification, considering the varying action dimensions across datasets (7 for LIBERO and ManiSkill, and 4 for MetaWorld), the invariance constraint is applied independently on each continuous action dimension, excluding the final binary gripper state. The threshold $\delta$ defaults to the dimension-specific standard deviation used for action denormalization in the original model checkpoint of the corresponding dataset. For efficient dual verification, we employ a W4A4-quantized action generator as the verification model. We set the denoising step of the verification model to 3 for GR00T and 5 for $\pi$-0 and $\pi$-0.5. The original action generator remains responsible for producing the candidate actions. All experiments are conducted on a single H200 GPU.

\noindent \textbf{Benchmarks.}
We evaluate A$^3$ on three representative embodied manipulation benchmarks: ManiSkill \cite{maniskill}, MetaWorld \cite{metaworld}, and LIBERO \cite{libero}. 
These benchmarks cover diverse multi-stage tasks requiring long-horizon planning, precise continuous control, and language-conditioned execution, providing a comprehensive testbed for adaptive execution commitment. We additionally conduct real-robot experiments to validate the practicality of A$^3$, including four tasks: \textit{FlipMug}, \textit{TapeBox}, \textit{HangMug}, and \textit{StackCube}. Specifically, we consider a tabletop manipulation setting equipped with an Agilex Piper robotic arm, where two Intel RealSense D435i cameras are deployed as the primary and wrist observations. Both A$^3$ and the baselines are instantiated using $\pi$-0.5, which is trained on a dataset that consists of four tasks, each with 50 expert demonstrations collected via teleoperation. During training, the horizon is set to $H=20$, and the control frequency of the robotic arm is fixed at 30 Hz. For evaluation, each result is obtained from 20 rollouts, except for the StackCube task, which is evaluated with 30 rollouts.

% • \textbf{ManiSkill} features compositional manipulation tasks with structured sub-goals, emphasizing long-horizon coordination across multiple stages. 
% Its staged task structure makes performance highly sensitive to execution commitment length, as over-committing unstable actions can cause cascading and irreversible failures.

% • \textbf{MetaWorld} consists of a diverse suite of continuous-control manipulation tasks with sparse rewards. 
% While individual tasks are shorter in horizon, multi-step open-loop execution leads to rapid error accumulation, making it well-suited for evaluating adaptive commitment strategies.

% • \textbf{LIBERO} is a large-scale language-conditioned benchmark covering four task suites: \emph{Spatial}, \emph{Object}, \emph{Goal}, and \emph{Long}. 
% These suites progressively increase task complexity and horizon length, from spatial reasoning to extended multi-stage manipulation. 
% The combination of visual-language grounding and long-horizon control introduces substantial prediction uncertainty, particularly in later rollout stages, making LIBERO a challenging setting for execution commitment.

\noindent \textbf{Metrics.}
We report \textbf{task success rate} as the primary performance metric.
To characterize execution behavior, we report the \textbf{average execution horizon length}, defined as the number of actions committed before each replan. This metric reflects how aggressively or conservatively a method commits predicted action chunks.
To measure computational cost, we report the \textbf{average number of replans}, defined as the total number of VLA forward passes invoked during an episode. %Together, execution horizon length and replans quantify the performance–efficiency trade-off of different commitment strategies.

\noindent \textbf{Baselines.}
We compare A$^3$ against three representative execution strategies. \textbf{Fixed Horizon.}
This baseline reflects the standard execution paradigm for VLA models, where a fixed number of actions $l \in \{1, \dots, H\}$ from each predicted chunk are executed before re-observing the environment.  
To provide a strong baseline, we determine the optimal fixed horizon via exhaustive grid search on the evaluation suite, selecting the $l$ that maximizes task success rate.
 \textbf{Mixture of Horizons (MoH).}
MoH \cite{moh} mitigates the trade-off between long-horizon foresight and short-term accuracy by decomposing the action chunk into segments with different effective horizons. 
Predictions from multiple horizon-specific segments are fused using a lightweight linear gating mechanism. 
Execution commitment is determined through cross-horizon consensus, forming a self-truncated executable chunk based on agreement among the different horizon voters.
\textbf{Adaptive Horizon.}
We additionally compare with recent adaptive execution methods based on different model-internal signals.
%AdaHorizon of EveryDayVLA \cite{everydayvla} estimates uncertainty from the discrepancy between concurrently generated continuous and discrete action predictions.
AutoHorizon \cite{autohorizon} uses self-attention weights as a proxy for the model's reliable prediction horizon.
%Adaptive Chunking \cite{so2026improving} measures the cosine similarity between the next action in the cached chunk and a newly predicted action, retaining the cached chunk when the two predictions remain consistent.

\subsection{Main Results}

\begin{table*}[t]
\caption{
Dynamic horizon comparison across LIBERO, MetaWorld, and ManiSkill.
We report average task success rate (Avg.) and average committed horizon length (Len.).
}
\centering
\small
\begin{adjustbox}{width=\textwidth}
\begin{tabular}{l c cccccc cc cc}
\toprule
\multirow{2}{*}{Backbone} & \multirow{2}{*}{Method} 
& \multicolumn{6}{c}{LIBERO} 
& \multicolumn{2}{c}{MetaWorld} 
& \multicolumn{2}{c}{ManiSkill} \\
\cmidrule(lr){3-8} \cmidrule(lr){9-10} \cmidrule(lr){11-12}
& 
& Spatial & Object & Goal & Long & Avg. & Len. 
& Avg. & Len. 
& Avg. & Len. \\
\midrule
\multirow{3}{*}{$\pi$-0}
& Original      & 97.2 & 98.2  & 95.6 & 84.6 & 95.1 & 5   & 78.0 & 3   & 77.6 & 5   \\ 
& MoH           & 97.6 & 98.8  & \textbf{96.4} & \textbf{87.4} & 95.1 & 5   & \textbf{79.4} & 3   & 78.0 & 5   \\
%& Adaptive Chunking           &   &     &   &  &   &      &    &     &   &    \\
& \textbf{A$^3$ (Ours)} & \textbf{98.2} & \textbf{99.0} & 96.2 & 87.2 & \textbf{95.2} & \textbf{9.4} & 79.2 & \textbf{3.2} & \textbf{78.5} & \textbf{6.1} \\
\midrule
\multirow{4}{*}{$\pi$-0.5}
& Original      & 98.6 & \textbf{99.8} & 97.6 & 95.4 & 97.9 & 6.3 & 77.8 & 3   & 88.1 & 5   \\ 
& MoH           & 98.4 & 99.0 & \textbf{98.2} & 95.2 & 97.7 & 5   & 78.4 & 3.3 & 88.4 & 5   \\
%& Adaptive Chunking           &   &     &   &  &   &      &    &     &   &    \\
%& EverydayVLA   & 98.2 & 99.2 & 97.4 & 95.6 & 97.6 & 6.8 & -    & -   & -    & -   \\
& AutoHorizon   & \textbf{99.1} & 99.2 & 97.5 & 91.6 & 96.9 & -   & -    & -   & -    & -   \\
& \textbf{A$^3$ (Ours)} & 98.6 & 99.3 & 98.2 & \textbf{96.0} & \textbf{98.0} & \textbf{9.8} & \textbf{79.4} & \textbf{4.5} & \textbf{89.0} & \textbf{5.2} \\
\midrule
\multirow{2}{*}{GR00T}
& Original      & 91.4 & 97.8 & 84.4 & 86.8 & 90.1 & 4.7 & - & - & - & - \\
& \textbf{A$^3$ (Ours)} & \textbf{93.8} & \textbf{99.2} & \textbf{87.2} & \textbf{91.4} & \textbf{92.9} & 4.5 & - & - & - & - \\
\bottomrule
\end{tabular}
\end{adjustbox}
\label{tab:main}
\end{table*}

\begin{table}[t]
\centering
\caption{Performance comparison on real-world tasks.}\label{tab real}
\adjustbox{max width=\textwidth}{%
\begin{tabular}{lcccccccc}
\toprule
\multirow{2}{*}{Setting} & \multirow{2}{*}{\shortstack{Exec\\Horizon}} 
    & \multicolumn{5}{c}{Success Rate (\%)} 
    & \multirow{2}{*}{\shortstack{Inference\\Calls}} 
    & \multirow{2}{*}{\shortstack{Total\\Steps}} \\
\cmidrule(lr){3-7}
 & & FlipMug & TapeBox & HangMug & StackCube & Average & & \\
\midrule
\multirow{4}{*}{Fixed horizon} 
      & 5  & 50 & 35  & 25 & 0    & 27.5 & 91.5 & 455.3 \\
      & 10 & 70 & \textbf{100} & 35 & 66.7 & 67.9 & 32.7 & 321.5 \\
      & 15 & 95 & \textbf{100} & 35 & \textbf{86.7} & 79.2 & 17.7 & 257.7 \\
      & 20 & 90 & 95  & 35 & 73.3 & 73.3 & 12.4 & 238.1 \\
\midrule
Ours  & 13.5 & \textbf{95} & \textbf{100} & \textbf{60} & 83.3 & \textbf{84.6} & 17.2 & 250.1 \\
\bottomrule
\end{tabular}%
}
\end{table}

\noindent \textbf{Comparison with baselines.}
Table~\ref{tab:main} shows that A$^3$ consistently maintains or improves task success while substantially extending the committed horizon. 
On $\pi$-0, A$^3$ matches the best LIBERO average success rate (95.1\%) but increases the execution length from 5 to 9.4 steps, and improves ManiSkill from 77.6\% to 78.6\%. 
On the stronger $\pi$-0.5, A$^3$ achieves the highest LIBERO average success (98.1\%) while extending the committed horizon from 6.3 to 9.7 steps, a 54\% increase in execution length with simultaneous performance gains. Meanwhile, our method achieves better performance efficiency tradeoff on Metaworld with 1.6\% improvement in success rate and future 1.5 horizon extension than manual choosing. 
%For GR00T, A$^3$ improves the LIBERO-Long suite from 86.8\% to 91.4\%, resulting in a 2.8\% gain in overall LIBERO average. 
These results demonstrate that our method enables longer yet reliable commitment, with execution horizons emerging adaptively from task difficulty.

\begin{figure}
    \centering
    \includegraphics[width=0.95\linewidth]{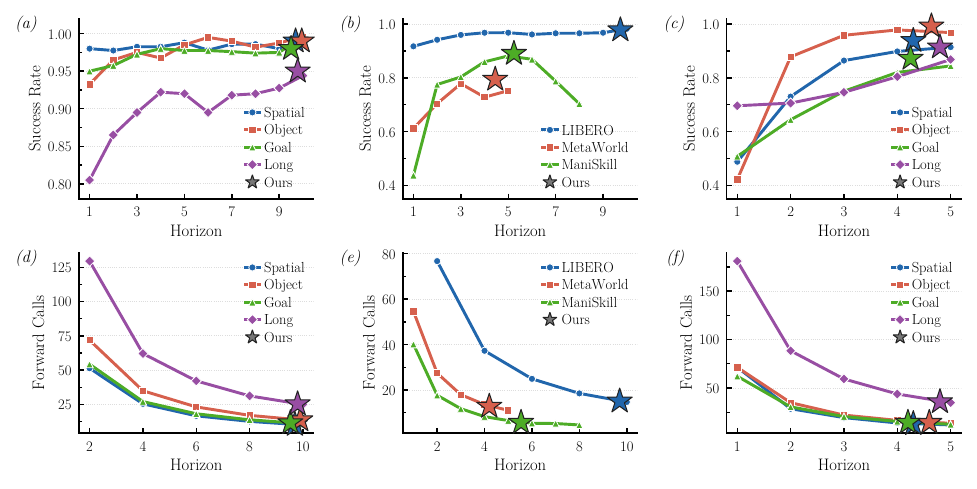}
    \caption{Trade-off between success rate (top row) and forward calls (bottom row) across execution horizons. The first two column experiments are conducted on $\pi$-0.5 while the last one is on GR00T.}
    \label{fig:tradeoff}
\end{figure}

\noindent \textbf{Efficiency--performance tradeoff.}
Figure~\ref{fig:tradeoff} illustrates the tradeoff between execution horizon,
success rate, and inference calls across task suites and benchmarks.
Longer horizons generally reduce forward calls but risk error accumulation,
leading to saturated or degraded success rates on challenging tasks.
Our method adaptively selects the execution horizon,
placing itself near the Pareto frontier: it achieves competitive or superior
success rates while requiring significantly fewer forward calls compared to
fixed short-horizon baselines, demonstrating that adaptive commitment
effectively balances performance and inference efficiency.

\textbf{Real-world results.} Table~\ref{tab real} summarizes the performance of $\text{A}^3$ 
on four physical manipulation tasks. $\text{A}^3$ achieves an average success rate of 84.6\% 
across FlipMug, TapeBox, HangMug, and StackCube, outperforming the best fixed-horizon baseline 
(79.2\% at horizon 15) while maintaining comparable inference calls (17.2 vs.\ 17.7) and total 
steps (250.1 vs.\ 257.7). The adaptive execution horizon converges to 13.5 steps on average 
without any task-specific tuning. Notably, the gains are most pronounced on contact-rich tasks: 
HangMug improves by 25\% over the best fixed horizon, as the adaptive mechanism shortens the 
committed prefix during the sub-centimeter alignment phase to maintain tighter visual feedback. 
In contrast, TapeBox achieves 100\% under both settings, suggesting that the task's relatively 
forgiving placement tolerance does not require fine-grained horizon adaptation. StackCube, which 
involves precise stacking under significant positional uncertainty, benefits most from shorter 
committed horizons during the critical placement phase, yielding a 16.6\% gain. These results 
demonstrate that $\text{A}^3$ naturally allocates control granularity according to task 
difficulty—committing longer horizons during free-space motion and shortening them at 
contact-critical moments—without any task-specific configuration.

\begin{table*}[t]
\centering
\caption{Performance comparison under different state perturbations using $\pi$-0.5 on LIBERO.
Masking indicates the fraction of pixels occluded. Blur uses Gaussian blur with kernel size $k$.}
\label{tab:perturbation_results}
\small
\setlength{\tabcolsep}{4pt}
\begin{adjustbox}{max width=\textwidth}
\begin{tabular}{l cc cccccc cccccc}
\toprule
& \multicolumn{2}{c}{Original}
& \multicolumn{6}{c}{Masking}
& \multicolumn{6}{c}{Blur} \\
\cmidrule(lr){2-3}
\cmidrule(lr){4-9}
\cmidrule(lr){10-15}
Method
& Avg $\uparrow$ & Len
& \multicolumn{2}{c}{50\%} & \multicolumn{2}{c}{55\%} & \multicolumn{2}{c}{60\%}
& \multicolumn{2}{c}{$k{=}11$} & \multicolumn{2}{c}{$k{=}13$} & \multicolumn{2}{c}{$k{=}15$} \\
\cmidrule(lr){4-5}\cmidrule(lr){6-7}\cmidrule(lr){8-9}
\cmidrule(lr){10-11}\cmidrule(lr){12-13}\cmidrule(lr){14-15}
& & 
& Avg $\uparrow$ & Len & Avg $\uparrow$ & Len & Avg $\uparrow$ & Len
& Avg $\uparrow$ & Len & Avg $\uparrow$ & Len & Avg $\uparrow$ & Len \\
\midrule
Original & 97.9 & 6.3 & 83.2 & 6.3 & 66.6 & 6.3 & 40.8 & 6.3 & 92.6 & 6.3 & 79.6 & 6.3 & 55.8 & 6.3 \\  
A$^3$ (Ours) & 98.1 & 9.7 & 89.0 & 8.9 & 72.6 & 7.5 & 51.0 & 6.3 & 96.8 & 9.6 & 89.6 & 8.9 & 66.0 & 8.1 \\ 
& \textcolor{green!60!black}{+0.2} & \textcolor{green!60!black}{+3.4}
& \textcolor{green!60!black}{+5.8} & \textcolor{green!60!black}{+2.6}
& \textcolor{green!60!black}{+6.0} & \textcolor{green!60!black}{+1.2}
& \textcolor{green!60!black}{+10.2} & \textcolor{green!60!black}{0}
& \textcolor{green!60!black}{+4.2} & \textcolor{green!60!black}{+3.3}
& \textcolor{green!60!black}{+10.0} & \textcolor{green!60!black}{+2.6}
& \textcolor{green!60!black}{+10.2} & \textcolor{green!60!black}{+1.8} \\
\bottomrule
\end{tabular}
\end{adjustbox}
\end{table*}

\noindent\textbf{Latency analysis.}
As shown in Table~\ref{tab:latency_breakdown}, A$^3$ incurs only 21.1\,ms of verification latency, substantially lower than AutoHorizon's 39.6\,ms. Although its per-call latency is slightly higher than the original policy, A$^3$ requires fewer inference calls than MoH and AutoHorizon, resulting in a lower episode latency of 2.9\,s versus 5.6\,s and 3.3\,s, respectively.

\begin{table}[t]
\centering
\small
\setlength{\tabcolsep}{4pt}
\caption{Inference latency at the call and episode levels. Per-call
latency (ms) is decomposed into VLM observation encoding, action-chunk
generation, and verification. Episode latency (s) denotes the accumulated
model inference time over an entire episode, excluding environment and
robot execution time. Experiments are conducted using $\pi$-0.5 with an H200 GPU on LIBERO-Spatial.}
\label{tab:latency_breakdown}
\begin{tabular}{l cccc cc}
\toprule
& \multicolumn{4}{c}{\textbf{Per inference call (ms)}} 
& \multicolumn{2}{c}{\textbf{Per episode}} \\
\cmidrule(lr){2-5}
\cmidrule(lr){6-7}
\textbf{Method}
& \textbf{Total}
& \textbf{VLM obs.}
& \textbf{Chunk gen.}
& \textbf{Verification}
& \textbf{\# Calls}
& \textbf{Latency (s)} \\
\midrule

Original
 & 253.2  & 90.9  & 162.3  &  --
& 10 & 2.6 \\
MoH
& 279.3  & 93.1  & 186.2  & -- 
& 20 &  5.6 \\

AutoHorizon
& 300.3  & 92.3  & 168.4  &  39.6 
& 11 & 3.3 \\

A$^3$ (ours)
& 289.5 & 93.3  &  175.1 & 21.1
& 10 & 2.9 \\
\bottomrule
\end{tabular}
\end{table}

\subsection{Robustness Analysis}

\noindent \textbf{State perturbations.}
We evaluate robustness under degraded observations by applying test-time \emph{Masking} (occluding 50/55/60\% of pixels) and \emph{Blur} (Gaussian blur with kernel size ($k$=11/13/15)); results are summarized in Table~\ref{tab:perturbation_results}. A$^3$ consistently improves success rate over the fixed-horizon baseline across all perturbations, with larger gains under stronger corruption (e.g., 10.2\% improvement under $k$=15 Gaussian blur). Meanwhile, A$^3$ generally achieves longer execution horizons, indicating more stable commitment under noisy states. Under the hardest 60\% masking, the committed horizon converges to 6.3 while still substantially increasing 10.2\% success rate. Overall, these results demonstrate that A$^3$ effectively mitigates error accumulation and remains reliable under realistic state degradations.

\begin{table}[t]
\centering
\small
\setlength{\tabcolsep}{5pt}
\caption{Effect of prediction horizon on execution horizon and LIBERO suites.}
\label{tab:horizon_tradeoff}
\begin{tabular}{c l c cccc}
\toprule
\textbf{Pred. Horizon} & \textbf{Method} & \textbf{Exec. Horizon} & \textbf{Spatial} & \textbf{Object} & \textbf{Goal} & \textbf{Long} \\
\midrule
\multirow{2}{*}{10} 
& Fixed horizon  &  10 & 98.6 & 99.6 & 97.6 &  94.6 \\
& Ours      & 9.7 & 98.6 & 99.4 & 98.0 & 96.2 \\  \midrule
\multirow{2}{*}{15} 
& Fixed horizon  & 15 & 91.4 & 96.4 & 96.4 & 89.6 \\
& Ours      & 14.2  & 97.2 & 98.6 & 96.8 &  92.6\\  \midrule
\multirow{2}{*}{20} 
& Fixed horizon  & 20 & 64.4 & 79.2 &  82.6 & 71.8 \\
& Ours      & 15.9 & 72.0 & 81.4  & 84.2 &  73.6 \\  
\bottomrule
\end{tabular}
\end{table}

\noindent \textbf{Effect of prediction chunk size.}
Table~\ref{tab:horizon_tradeoff} examines how prediction chunk size affects
execution horizon and task performance on LIBERO suites.
At a chunk size of 10, both methods perform comparably, with our method
showing a marginal improvement on Long tasks.
As the chunk size increases, the fixed-horizon baseline degrades substantially,
most notably on Spatial, where the success rate drops from 98.6 to 64.4.
Our method consistently mitigates this degradation by adaptively reducing the committed execution horizon: at chunk sizes of 15 and 20, we outperform
the fixed baseline across all four task suites, with the largest gains on
Spatial (5.8\% and 7.6\% respectively).

\section{Ablation Study}
\label{appendix}

\noindent\textbf{Effect of components.}
Table~\ref{tab:ablation_modules} presents a staged ablation of the two
verification components on GR00T-LIBERO, $\pi_{0.5}$-ManiSkill, and
$\pi_0$-MetaWorld. Consensus scoring is not evaluated as a standalone
component because it serves as an intermediate statistic for selecting the
dominant-mode medoid and ordering the conditional verification of individual
actions. Starting from the original policy, introducing
\emph{consensus-ordered conditional invariance} improves success across all
three benchmarks (90.1$\rightarrow$92.1, 88.1$\rightarrow$88.3, and
78.0$\rightarrow$78.6) while extending the average execution horizon,
particularly on ManiSkill and MetaWorld. This indicates that anchoring
re-decoding on higher-consensus actions enables the policy to make longer
commitments while filtering conditionally unstable predictions.
Adding \emph{prefix-closed sequential consistency} further improves success
(92.1$\rightarrow$92.9, 88.3$\rightarrow$89.1, and
78.6$\rightarrow$79.2), while moderately shortening the execution horizon
on all three benchmarks. This behavior is consistent with its role of
truncating execution at the first verification failure, thereby preventing
over-commitment to later actions. Overall, consensus-ordered invariance
enables adaptive commitment, whereas prefix-closed consistency provides a
complementary safeguard that favors shorter but more reliable prefixes.

\begin{table}[h]\caption{Ablation study of each component on three benchmarks. We report the success rate/execution horizon length for each setting.}
\centering
\small
\setlength{\tabcolsep}{6pt}
\begin{tabular}{lccc}
\toprule
\textbf{Module} & \textbf{GR00T-LIBERO} & \textbf{$\pi$0.5-ManiSkill} & \textbf{$\pi$0-MetaWorld} \\
\midrule
Baseline (original)                  & 90.1/4.70  & 88.1/5.00 & 78.0/3.00 \\
% + Consensus scoring                         & 90.6/4.70 & 88.1/5.00 & 78.2/3.00 \\
+ Consensus-ordered conditional invariance           & 92.1/4.74 & 88.3/5.77 & 78.6/3.41 \\
+ Prefix-closed sequential consistency            & 92.9/4.52  & 89.1/5.24 & 79.2/3.23 \\
\bottomrule
\end{tabular}
\label{tab:ablation_modules}
\end{table}

\noindent \textbf{Parameter sensitivity analysis.} We analyze the two key hyperparameters of $\text{A}^3$ on $\pi$-0.5 under LIBERO in Table~\ref{tab param}. For the sample count $K$, success rate saturates quickly, with $K=8$ recovering most gains over $K=4$ while adding only 18.4ms overhead; we adopt $K=8$ as default. For the acceptance threshold $\delta$, both success rate and average horizon remain stable across $\delta \in \{0.5, 1.0, 1.5\} \times \textit{std}$, confirming that setting $\delta$ to the dimension-specific denormalization standard deviation provides a natural, dataset-adaptive calibration without manual tuning.

\begin{table}[h]\caption{Parameter sensitivity of $K$ and $\delta$ on $\pi$-0.5 under LIBERO. Here, ``std" denotes the dimension-specific standard deviation used for action denormalization in the original model checkpoint of the corresponding dataset.}
\centering
\begin{tabular}{ccc|ccc}
\toprule
parameter & success rate & latency(ms) & parameter & success rate & average horizon \\ \midrule
   K = 4       &   97.3           &   467.2      &  $\delta$ = 0.5 * std       &     97.9         &    9.6             \\
   K = 8       &   98.1           &  485.6       &     $\delta$ = 1 * std      &    98.1          &      9.7          \\ 
   K = 16       &      98.2        &  503.6       &   $\delta$ = 1.5 * std        &   97.8           &       9.8         \\    
\bottomrule
\end{tabular}\label{tab param}
\end{table}

\begin{figure*}[t]
    \centering
    \includegraphics[width=\linewidth]{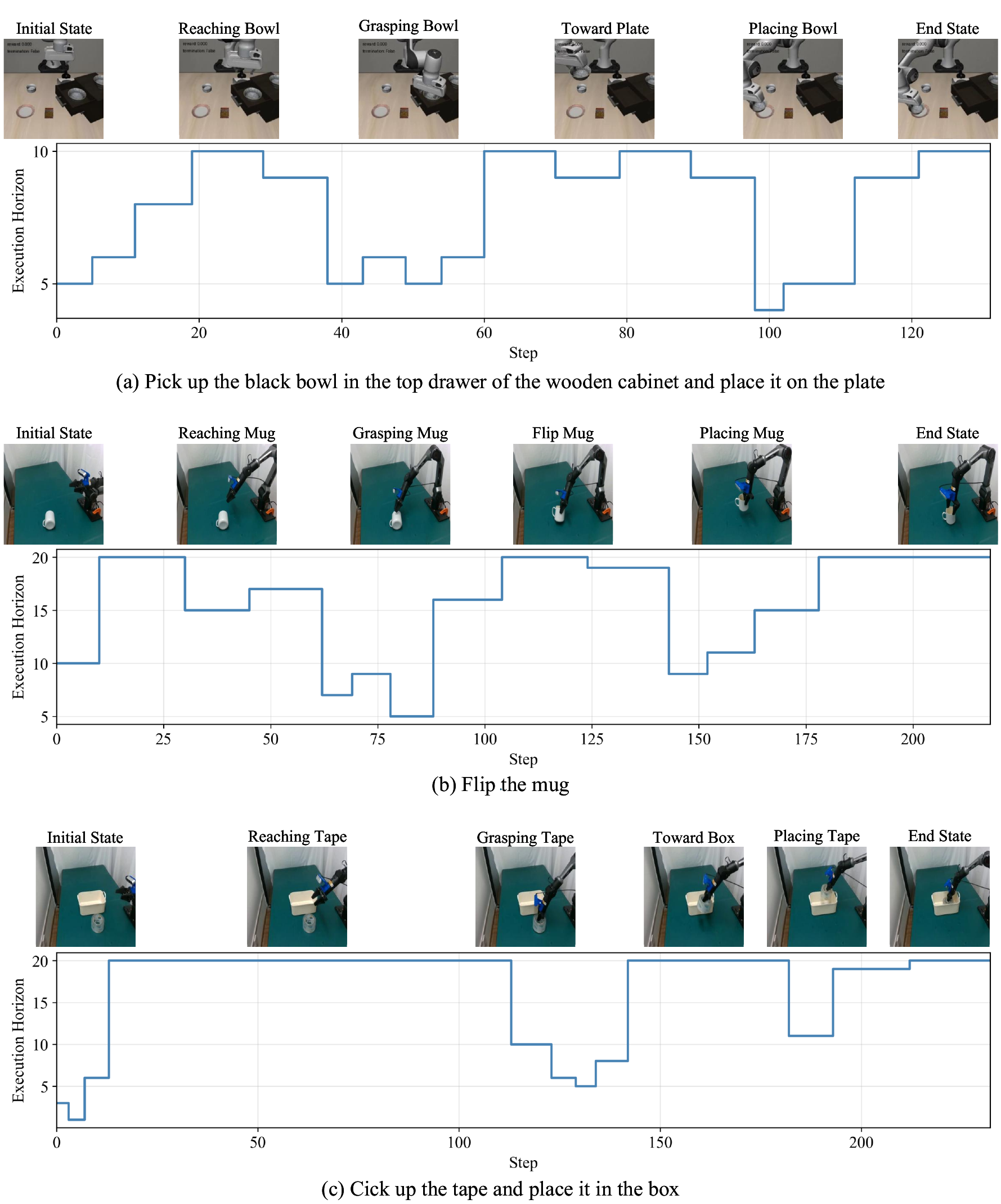}
    \caption{Visualization of the execution horizon across different tasks.}
    \label{fig:vis}
\end{figure*}

% \noindent\textbf{Visualization analysis.}
\noindent \textbf{Visualization of adaptive horizon behavior.} Figure~\ref{fig:vis} visualizes how the execution horizon evolves throughout the rollout on three representative tasks. A consistent pattern emerges across all of them. During free-space motion, such as approaching the target object or transferring between locations, the policy adopts a relatively large horizon to accelerate execution. In contrast, during contact-rich phases such as grasping, placing, and flipping, the horizon automatically shrinks to ensure sufficient feedback frequency for fine-grained control.

For the task of picking up the black bowl from the top drawer of the wooden cabinet and placing it on the plate, shown in Figure ~\ref{fig:vis}(a), the rollout involves reaching into a confined space, grasping the bowl, transferring it out of the drawer, and finally placing it on the plate, leading to frequent alternation between long and short horizon segments. The flip-the-mug task in Figure ~\ref{fig:vis}(b) requires the gripper to grasp the rim of an inverted mug, lift and rotate it upright, and then release it, where free-space transitions are executed with large horizons while the grasping and uprighting phases shrink to smaller values to handle the precise alignment with the thin circular rim and the controlled rotation. The pick-and-place tape task in Figure ~\ref{fig:vis}(c) follows a more standard pattern, in which the horizon shrinks only at the two critical moments of grasping and releasing, while remaining large during the rest of the trajectory, leading to a more compact overall execution.

These comparisons demonstrate that the proposed adaptive horizon mechanism is able to allocate control granularity according to the underlying task structure, executing efficiently in simple settings and operating with finer precision in more complex ones, all without task-specific tuning.

\begin{figure}[t]
    \centering
    \includegraphics[width=\linewidth]{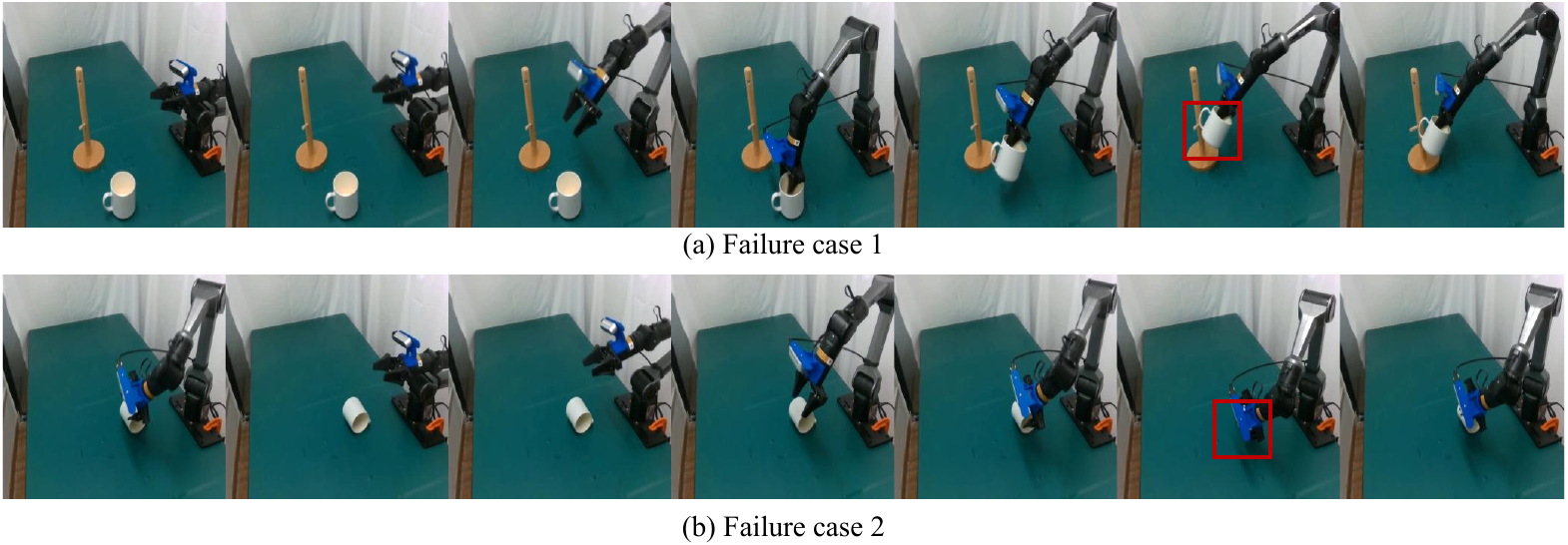}
    \caption{Two representative failure cases. \textbf{(a)} Misalignment between the mug handle and the hook in the \textit{hang mug} task. \textbf{(b)} Self-occlusion of the inverted mug rim by the gripper in the \textit{flip mug} task.}
    \label{fig:failure}
\end{figure}

\textbf{Failure case analysis.} 
We analyze two representative failure cases from real-world rollouts. As shown in Figure ~\ref{fig:failure}(a), in the \textit{hangmug} task, the mug handle fails to align with the hook during the placing phase. 
This task requires sub-centimeter alignment, and small residual errors 
in the predicted end-effector pose, though tolerable earlier in the 
trajectory, become unrecoverable at the moment of contact. $\text{A}^3$ 
improves the reliability of committed action prefixes but does not 
enhance the spatial precision of the underlying policy, and is 
therefore insufficient to resolve such fine-grained contact-rich 
failures.

As shown in Figure ~\ref{fig:failure}(b), in the \textit{flipmug} task, 
the gripper completely occludes the inverted mug rim from the top-down 
camera view during approach, leaving the policy without effective 
visual feedback at the critical closing moment and ultimately leading 
to a failed grasp. Since $\text{A}^3$ verifies action consistency 
under the model's existing observations, it cannot compensate for 
missing visual information caused by viewpoint occlusion.
Overall, these cases indicate that adaptive execution commitment alone 
is insufficient when failures originate from perceptual coverage or 
control precision limits of the underlying policy.

%\subsection{Ablation Study}

% \noindent \textbf{Ablation about each component, parameter sensitivity, latency breakdown, visualization of adaptive horizon behavior, and failure case analysis can be found in Sec \ref{appendix}.}

\section{Conclusion and Future Work}

In this work, we identify the determination of execution commitment as a principled yet underexplored inference problem in multi-step VLA systems, and reformulate horizon selection as a state-conditioned prefix verification problem. Instead of relying on predefined execution windows or heuristic uncertainty thresholds, $\text{A}^3$ treats predicted action chunks as speculative drafts and adaptively determines the executable horizon through structured self-verification, without auxiliary modules or task-specific retraining. By combining trajectory-level consensus estimation with dual hierarchical verification constraints, \textit{i.e.,} consensus-ordered conditional invariance and prefix-closed sequential consistency, $\text{A}^3$ grounds execution decisions in both dominant sampled
plan and sequential execution consistency. Extensive experiments across diverse VLA backbones and benchmarks demonstrate that $\text{A}^3$ eliminates the need for manual horizon tuning while consistently matching or outperforming exhaustively grid-searched fixed horizons, achieving a superior trade-off between execution robustness and inference efficiency.

\textbf{Future Work.}
While A$^3$ demonstrates the viability of self-speculative prefix
verification for adaptive execution commitment, several limitations motivate
further investigation. First, replacing group sampling with lightweight
learned reliability estimators could reduce inference overhead while
preserving horizon-selection performance. Second, because all verification
signals are derived from the underlying policy, A$^3$ cannot identify
confidently incorrect predictions that remain stable across sampling and
structured re-decoding. Incorporating explicit world models or independently
learned state-transition predictors could complement self-consistency with
external evidence and improve the detection of systematic errors,
particularly under distribution shift. Finally, prefix-closed acceptance
conservatively discards the entire suffix following the first verification
failure. Selectively re-planning and re-verifying such suffixes could recover
usable subsequent actions while preserving causal execution.

\bibliographystyle{unsrt}
\bibliography{main}

\appendix

\begin{figure}
    \centering
    \includegraphics[width=1\linewidth]{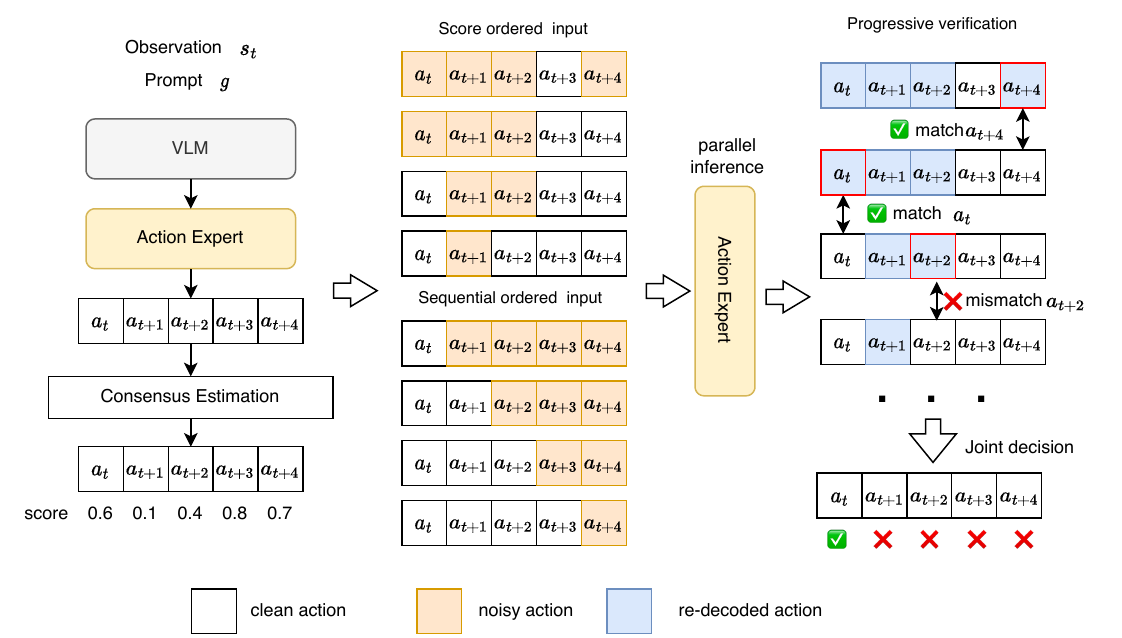}
    \caption{Implementation of dual verification tree.}
    \label{fig:main}
\end{figure}

\section{Implementation Details}
\label{sec detail}
\noindent\textbf{Implementation of the verification tree.}
As shown in Figure \ref{fig:main}, following the self-speculative decoding paradigm of SSD~\cite{ssd}, we implement the 
dual hierarchical verification as a batched tree over candidate action prefixes. 
For a prediction horizon of $H$ actions, the two verification constraints—consensus-ordered 
conditional invariance and prefix-closed sequential consistency—each requires $H$ re-decoding 
queries, yielding $2H$ total inpainting inputs. Rather than executing these sequentially, 
we construct all $2H$ queries in parallel: for each action position $i$, the corresponding 
context actions (either $\mathcal{C}^{\mathrm{inv}}_i$ or $\mathcal{C}^{\mathrm{seq}}_i$) 
are fixed at their draft values via flow matching inpainting, while position $i$ is freely 
denoised. All $2H$ inpainting inputs are stacked into a single batch and processed in one 
forward pass through the action expert, after which the re-decoded outputs are compared 
against the original draft using the acceptance threshold $\delta$. This design ensures 
that the entire verification stage adds exactly one forward call regardless of horizon 
length $H$, keeping the per-step overhead constant and amortizable over the committed 
action prefix.

\noindent \textbf{Algorithm.} The algorithm of our method can be found in Alg. \ref{alg:a3}.

\begin{algorithm}[t]
\caption{Adaptive Action Acceptance ($\text{A}^3$)}
\label{alg:a3}
\begin{algorithmic}[1]
\Require VLA policy $\pi_\theta$, current state $s_t$, prediction horizon $H$,
         number of samples $K$, acceptance threshold $\delta$, alignment window $w$,
         temperature $\tau$, kinematic integrator $\mathcal{G}$
\Ensure Committed action prefix $\mathbf{A}_t[1:l^*]$

\vspace{2pt}
\Statex \textbf{Stage 1: Mode-aware Trajectory Consensus Scoring}
\vspace{2pt}

\State Generate $K$ candidate chunks via parallel noise injection:
       $\{\mathbf{A}_t^{(k)}\}_{k=1}^K \leftarrow \pi_\theta(s_t,\, \epsilon^{(k)})$,\quad $\epsilon^{(k)} \sim \mathcal{N}(0, \sigma^2 I)$
\State Compute induced trajectories $\{T^{(k)}\}_{k=1}^K$ via kinematic integrator $\mathcal{G}$
\For{each pair $(k, k')$}
    \State $D(k,k') = \sum\nolimits_j \min_{|\Delta| \leq w} \|\hat{s}^{(k)}_{t+j} - \hat{s}^{(k')}_{t+j+\Delta}\|_2^2$
\EndFor
\State Cluster $\{T^{(k)}\}$ under $D(\cdot,\cdot)$;\quad score each cluster $c$: $S_c = p_c \cdot \exp\!\left(-\mathrm{Disp}(c)/\tau\right)$
\State $c^* \leftarrow \arg\max_c S_c$;\quad
       $k^* \leftarrow \arg\min_{k \in c^*} \sum_{k' \in c^*} D(k,k')$ \Comment{dominant mode \& medoid draft}
\State Compute per-step consensus score $\tilde{R}_j$ for $a_{t+j} \in \mathbf{A}_t^{(k^*)}$ via Eq.~(5)

\vspace{2pt}
\Statex \textbf{Stage 2: Dual Hierarchical Verification via Flow Matching Inpainting}
\vspace{2pt}

\State Sort action indices by consensus (descending): $\pi_R \leftarrow \mathrm{argsort}(\{\tilde{R}_j\}_{j=1}^H,\downarrow)$
\State Construct batched inpainting inputs $\mathcal{B} \leftarrow \emptyset$
\For{each action index $i \in \{1,\ldots,H\}$}
    \State $\mathcal{C}^{\mathrm{inv}}_i \leftarrow \{j \mid \tilde{R}_j > \tilde{R}_i\}$ \Comment{higher-consensus positions as fixed context}
    \State $\mathcal{C}^{\mathrm{seq}}_i \leftarrow \{j \mid j < i\}$ \Comment{sequential prefix positions as fixed context}
    \State Construct inpainting mask $m^{\mathrm{inv}}_i$: fix positions $\mathcal{C}^{\mathrm{inv}}_i$ to $\mathbf{A}_t^{(k^*)}$, denoise position $i$
    \State Construct inpainting mask $m^{\mathrm{seq}}_i$: fix positions $\mathcal{C}^{\mathrm{seq}}_i$ to $\mathbf{A}_t^{(k^*)}$, denoise position $i$
    \State $\mathcal{B} \leftarrow \mathcal{B} \cup \{(m^{\mathrm{inv}}_i,\, m^{\mathrm{seq}}_i)\}$
\EndFor
\State Execute single batched flow matching denoising pass over $\mathcal{B}$:
       $\{\tilde{a}^{\mathrm{inv}}_i,\, \tilde{a}^{\mathrm{seq}}_i\}_{i=1}^H \leftarrow \pi_\theta(s_t \mid \mathcal{B})$
\For{each action index $i \in \{1,\ldots,H\}$}
    \State $V^{\mathrm{inv}}_i \leftarrow \mathbf{1}\!\left[\,|\tilde{a}^{\mathrm{inv}}_i - a_{t+i}| < \delta\,\right]$ \Comment{conditional invariance check}
    \State $V^{\mathrm{seq}}_i \leftarrow \mathbf{1}\!\left[\,|\tilde{a}^{\mathrm{seq}}_i - a_{t+i}| < \delta\,\right]$ \Comment{sequential consistency check}
\EndFor

\vspace{2pt}
\Statex \textbf{Stage 3: Prefix-closed Horizon Determination}
\vspace{2pt}

\State $l^* \leftarrow \max\,\{l \in \{1,\ldots,H\} \mid \forall\,i \leq l:\ V^{\mathrm{inv}}_i = 1\ \wedge\ V^{\mathrm{seq}}_i = 1\}$
\State \Return $\mathbf{A}_t[1:l^*] = \{a_t,\ldots,a_{t+l^*-1}\}$

\end{algorithmic}
\end{algorithm}

% \newpage
% \input{checklist.tex}

\end{document}